# Interpret Federated Learning with Shapley Values

**Guan Wang**
Digital and Smart Analytics, Swiss Re, Hong Kong
guan_wang@swissre.com

## Abstract

Federated Learning is introduced to protect privacy by distributing training data into multiple parties. Each party trains its own model and a meta-model is constructed from the sub models. In this way the details of the data are not disclosed in between each party. In this paper we investigate the model interpretation methods for Federated Learning, specifically on the measurement of feature importance of vertical Federated Learning where feature space of the data is divided into two parties, namely host and guest. For host party to interpret a single prediction of vertical Federated Learning model, the interpretation results, namely the feature importance, are very likely to reveal the protected data from guest party. We propose a method to balance the model interpretability and data privacy in vertical Federated Learning by using Shapley values to reveal detailed feature importance for host features and a unified importance value for federated guest features. Our experiments indicate robust and informative results for interpreting Federated Learning models.

## 1 Introduction

Federated Learning or Federated Machine Learning (FML) [Jakub et al., 2016] is introduced to solve privacy issues in machine learning using data from multiple parties. Instead of transferring data directly into a centralized data warehouse for building machine learning models, Federated Learning allows each party to own the data in its own place and still enables all parties to build a machine learning model together. This is achieved either by building a meta-model from the sub-models each party builds so that only model parameters are transferred, or by using encryption techniques to allow safe communications in between different parties [Qiang et al., 2019].

Federated Learning opens new opportunities for many industry applications. Companies have been having big concerns on the protection of their own data and are unwilling to share with other entities. With Federated Learning, companies can build models together without disclosing their data and share the benefit of machine learning. An example of Federated Learning use case in insurance industry is shown in Figure 1.

Primary insurers, reinsurers and third-party companies like online retailers can all work together to build machine learning models for insurance applications. Number of training instances is increased by different insurers and reinsurers, and feature space for insurance users is extended by third-party companies. With the help of Federated Learning, machine learning can cover more business cases and perform better.

However, industries like insurance or banking are heavily regulated. Models we build are supposed to be audited and understood by compliance and legal parties. Model interpretation has been an active research area in recent years [Scott et a. 2017] [Ribeiro et al., 2016] [Shrikumar et al., 2017] [Datta et al., 2016] [Bach et al., 2015] [Lipovetsky et al., 2001]. Some techniques are model dependent especially for linear models and decision trees, while some are model agnostic that can be applied to any supervised machine learning model. A good review of interpretable machine learning can be found in a recent book [Molnar et al., 2019].

There are still challenges specifically to Federated Learning models as partial data are private and cannot be read or analyze. For single prediction interpretation for vertical Federated Learning, interpretation results or the feature importance can reveal the underlined feature data from the other parties. Interpreting models while not breaking federated privacy is critical for FML to be used in production.

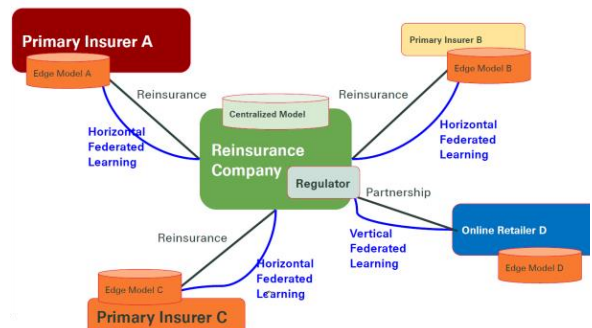

Figure 1: Federated Learning for Insurance Industry. Horizontal FML corresponds to primary insurers working with a reinsurer (same feature space). Vertical FML corresponds to reinsurer working with another data provider like Online Retailer (extended feature space).

In this paper we propose a method to use Shapley value to compromise the model interpretability and privacy protection in vertical FML by revealing detailed feature importance for owned features and a unified feature importance for the features from the other parties. Our experiment results indicate robust and informative results for interpreting Federated Learning models.

In the next chapters of this paper, we first briefly introduce Federated Learning and Model Interpretation separately. We then cover the Shap Federated method we propose on interpreting FML models, followed by some experiments. We conclude the paper with some discussions in the last chapter.

## 2 Federated Learning

Federated Learning originated from some academic papers like [Jakub et al., 2016] [H. Brendan et al., 2016] and a follow-up blog[1] from Google in 2017. The idea is that Google wants to train its own input method on its Android phones called "Gboard" but does not want to upload the sensitive keyboard data from their users to Google's own servers. Rather than uploading user's data and training models in the cloud, Google lets users train a separate model on their own smartphones (thanks to the neural engines from several chip manufacturers) and upload those black-box model parameters from each of their users to the cloud and merge the models, update the official centralized model, and push the model back to Google users. This not only avoids the transmission and storage of user's sensitive personal data, but also utilizes the computational power on the smartphones (a.k.a the concept of Edge Computing) and reduce the computation pressure from their centralized servers.

When the concept of Federated Learning was published, Google's focus was on the transmission of models as the upload bandwidth of mobile phones is usually very limited. One possible reason is that similar engineering ideas have been discussed intensively in distributed machine learning. The focus of Federated Learning was thus more on the "engineering work" with no rigorous distributed computing environment, limited upload bandwidth and slave nodes as massive number of users.

Data privacy is becoming an important issue, and lots of relating regulations and laws have been taken into action by authorities and governments [Albrecht 2016] [Goodman and Flaxman 2016]. The companies that have been accumulating tons of data and have just started to make value of it now have their hands tightened. On the other hand, all companies value a lot their own data and feel reluctant from sharing data with others. Information islands kill the possibility of cooperation and mutual benefit. People are looking for a way to break such prisoner dilemma while complying with all the regulations. Federated Learning was soon recognized as a great solution for encouraging collaboration while respecting data privacy.

---

[1] "Federated Learning: Collaborative Machine Learning without Centralized Training Data", https://ai.google-blog.com/2017/04/federated-learning-collaborative.html

**Categorization of Federated Learning**
[Qiang et al., 2019] describes Federated Learning in three categories: Horizontal Federated Learning, Vertical Federated Learning and Federated Transfer Learning. Such categorization extends the concept of Federated Learning and clarify the specific solutions under different use cases.

*Horizontal Federated Learning* applies to circumstances where we have a lot of overlap on features but only a few on instances. This refers to the Google Gboard use case and models can be ensembled directly from the edge models.

*Vertical Federated Learning* refers to where we have many overlapped instances but few overlapped features. An example is between insurers and online retailers. They both have lots of overlapped users, but each owns their own feature space and labels. Vertical Federated Learning merges the features together to create a larger feature space for machine learning tasks and uses homomorphic encryption to provide protection on data privacy for involved parties.

*Federated Transfer Learning* uses Transfer Learning [Sinno et al. 2010] to improve model performance when we have neither much overlap on features nor on instances.

We give an example in insurance industry in Figure 1 to illustrate the idea. Horizontal FML corresponds to primary insurers working with a reinsurer. For the same product primary insurers share the similar features. Vertical FML corresponds to reinsurer working with another third-party data provider like online retailer. An online retailer will have more features for a certain policyholder that can increase the prediction power for models built for insurance.

For a detailed introduction of Federated Learning and their respective technology that is used, please refer to [Qiang et al., 2019].

## 3 Model Interpretation

For machine learning to work in industrial applications, model interpretation is very important especially from regulatory and legal perspective. Audition is needed before a machine learning model is deployed for production to avoid unethical cases like discrimination.

**Categorization of Interpretation Methods**
How to interpret machine learning models is an active research topic. In our opinion model interpretation methods can be put into 3 categories.

First category is *Interpretation on Results*. It is the easiest and simplest method to directly analyze the prediction results from the machine learning model with some simple data mining and visualization methods. Cases like discrimination on gender are easy to spot. This method will naturally work for Horizontal FML. For vertical FML since the host does not know about the features from guests, statistical analysis on features is not likely to work unless we design some secure mechanism for sharing statistics [Rivest et al. 1978].

*Interpretation on Model* is another interesting idea. Rule-based systems like decision trees or linear models are already possible to interpret by themselves. For rule systems we can trace back to the decision path for each prediction. For linear models we can check the linear weight of each feature. For non-linear models, directly interpreting the model can be difficult. In efforts like [N Frosst et al. 2017] researchers are trying to approximate complex models like Deep Neural Networks with simple models like decision trees so that the decision logic is easier to understand.

Finally, *Interpretation on Features* is an active research topic, mainly on model-dependent and model-agnostic techniques for estimating feature importance for each prediction. For example, LIME [Ribeiro et al., 2016] tries to use linear model to fit the original model locally for certain prediction results to reveal the feature importance, while SHAP [Scott et a. 2017] uses Shapley value to experiment on original models with different feature combinations to get the feature importance. This method also naturally works for horizontal FML.

In the next chapter we focus on making interpretations on features for vertical FML. Before that we first describe how we can use Shapley values to interpret general machine learning models.

**Shapley Values for Model Interpretation**

Shapley Value of one specific feature is the average marginal contribution of this feature across all possible feature combinations. We define Shapley value of a feature as [Molnar et al., 2019]:

$$\phi_j(val) = \sum_{S \subseteq \{x_1,\ldots,x_p\} \setminus \{x_j\}} \frac{|S|!\,(p-|S|-1)!}{p!} \left(val(S \cup \{x_j\}) - val(S)\right)$$

where S is a subset of the features used in the model, x is the vector of feature values of the instance to be explained and p the number of features. $val_x(S)$ is the prediction for feature values in set S that are marginalized over features that are not included in set S:

$$val_x(S) = \int \hat{f}(x_1,\ldots,x_p) dP_{x \notin S} - E_X\left(\hat{f}(X)\right)$$

For the target feature we are evaluating, we try all the combinations of the other features with this certain feature turned on and off and evaluate the difference to the original prediction results as the marginal contribution to this target feature.

---

**Algorithm 1** Calculate Shapley value for feature j

**Input**: x, instance of interest,
**Input**: f, machine learning model
**Input**: S, all possible combinations of features other than j
**Output**: $\phi_j(x)$ Shapley value for feature j
1: **for** each s in S **do**
2:     Construct instance as $x_{+j}$ = x [j is on, s]
3:     Construct instance as $x_{-j}$ = x [j is off, s]
4:     Calculate marginal contribution as $\phi_j^s = f(x_{+j}) - f(x_{-j})$
5: **end for**
6: Compute Shapley value as the average: $\phi_j(x) = \frac{1}{len(S)} \phi_j^s$

---

More specifically, we calculate Shapley value for each feature in Algorithm 1.

## 4 Interpreting Federated Learning Models

Most model interpretation methods can be directly used for Horizontal Federated Learning as all parties have data for the full feature space. There is no special issue for interpreting prediction results on both training data and new data, for both specific single predictions as granular check or for batch predictions as holistic check.

Vertical Federated Learning raises new issues for Model Interpretation where the feature space is divided into different parties. Directly using methods like Shapley values for each prediction will very likely reveal the protected feature value from the other parties and cause privacy issues. It is not trivial to develop a safe mechanism for vertical Federated Learning and find a balance between model interpretation and data privacy.

We propose a variant version of SHAP [Scott et a. 2017] to use Shapley value for FML model interpretation. We take dual-party (Host and Guest) vertical Federated Learning as an example, but the idea can be extended to multiple parties. Host owns the label data $y_i$ and part of the feature space $x_i^H$. Guest owns another part of the feature space $x_i^G$. Here i=1,…,n as we suppose host and guest have n overlapped instances with the same IDs. By using vertical FML, host and guest collaborate to develop a machine learning model for predicting Y. Now host wants to interpret a specific

---

**Algorithm 2** Shap Federated for calculating Feature Importance for Vertical FML

**Input**: x, instance of interest,
**Input**: f, machine learning model
**Input**: S, all possible combinations of {host features, $j^{fed}$}, where $j^{fed}$ is the united federated feature
**Output**: Feature Importance of Host Feature and Federated Feature
1: **for** each s in S **do**
2:     Host sets x' with the corresponding host features in s to be original value of x
3:     Host sets x' with the other host features to be reference value
4:     **if** index of $j^{fed}$ is in s **then**
5:         Host sends encrypted ID of x to guest
6:         Guest sets x' with all guest features to be original value of x
7:     **else**
8:         Host sends special ID to guest
9:         Guest sets all guest feature to be reference value
10:    **end if**
11:    Guest and Host run federated model prediction for x'
12:    Host stores prediction result of current feature combination s
13: **end for**
14: Feature Importance = Shapley values from all prediction results (with Algorithm 1)
15: **return** Feature Importance

prediction the model makes for instance x by looking at the Shapley values of the features. Instead of guest giving out the feature importance for all its feature space $x^G$, we combine the feature space of $x^G$ as *one united federated feature* $x^{fed}$, and compute the Shapley value for each of the host features $x^H$ and this federated feature $x^{fed}$.

Since this method requires to turn on and off certain features for calculating the Shapley value, for the federated feature host will need a specific ID to inform guest to return its part of the prediction with all its features turned off. For models that takes in NA values, this mean that the features will be set to NA. For models that cannot handle missing values, we follow the practice of [Scott et a. 2017] to set the feature values to be the median value of all the instances as the reference value.

In summary, this process can be described in Algorithm 2.

## 5. Experiments

We develop the Shap Federated algorithm based on [Scott et a. 2017].[2] We test our algorithm on the Standard Adult Census Income dataset [Ron et al., 1996]. This dataset is used to predict whether an individual's income exceeds $50K/yr based on census data. Our data contains 26048 instances with 12 attributes as features, as shown in Table 1.

We normalize the data and used Scikit-learn [Pedregosa et al. 2011] to train a KNN (K-Nearest-Neighbor) model for the income classification task. Number of neighbors is set to be 5. Algorithm is set to be KDTree to be used for computing the nearest neighbors. Leaf size for KDTree is set to be 30. The power parameter for the Minkowski metric is set to be 2. We randomize the data and use 80% instances for training and 20% for testing. For testing data the accuracy is 77.34%. We choose three work-related features as the guest features, namely *Working class, Occupation* and *Hours per Week*. Those three features are supposed to be provided by guest, and host cannot see the true values of them for every instance. The three guest features are united as one federated feature for model interpretation. In another experiment we also add two more features and try with five guest features of *Working class, Occupation, Hours per Week, Capital Gain* and *Capital Loss*.

We pick one specific instance in the data and test our Shap Federated algorithm to interpret the prediction results in the host feature space and the unified federated feature from guest. We also use the non-federated Shap algorithm to test on the whole feature space combined with both host and guest features as comparison. The example result is shown in Figure 2.

We then randomly sample 1000 instances and run the same experiment with Shap Federated algorithm for whole feature space, three federated guest features and five federated guest features. The scatter plot is in Figure 3. We take the mean Shapley value and the bar plot is in Figure 4.

Some observations are below:
- With some features federated into one unified feature, the Shap Federated algorithm can still get very close result of host feature importance compared to the result of using the whole feature space. Exact value and relative ranking of the feature importance is preserved well.
- The unified federated feature gives representative importance for the guest features that are hidden behind it.
- With more features being federated, Shap Federated algorithm can still give accurate feature importance results and is quite robust.

## 6. Conclusion

Data privacy is the major concern for Federated Learning and raises challenge for model interpretation for FML models. In this paper. we proposed that we can use Shapley values to interpret Federated Learning models especially vertical FML models without sacrificing data privacy. Our experiment results indicate robust host feature importance results for partial feature space compared to results of the whole feature space. Moreover, the Shapley value of the unified federated feature gives a reasonable indication of the holistic contribution of the federated features from the guest party while the guest data is totally hidden from the host view.

In real applications, our work not only enables us to interpret the model prediction, but also help us understand and quantify the contribution from the guest features, without knowing the detailed values of guest data. When guest data is introduced, in addition to measurement of the model performance, Shapley values of the united federated feature give another method to quantify the value of guest data for building the model.

Our work for interpreting FML model is also model agonistic, meaning that this should work for almost any kind of machine learning algorithms. There are also many other ways to interpret models, some are model dependent like tree interpreter in [Saabas et al.], some are also model agnostic. Our paper is the first trial to extend those model interpretation techniques to Federated Learning.

We expect our work can be built into FML toolsets like FATE[3] and TFF[4] and become the start of developing a standard model interpretation module for Federated Learning that is critical for industrial applications.

---

[2] Code can be found at https://github.com/crownpku/federated_shap

[3] FATE: Federated AI Technology Enabler, https://github.com/WeBankFinTech/FATE

[4] TensorFlow Federated: Machine Learning on Decentralized Data, https://www.tensorflow.org/federated

| Role | Attribute | Content |
| --- | --- | --- |
| *Label* | *Income* | >50K, <=50K. |
| Host Features | Age | Continuous. |
| | Country | United-States, Cambodia, England, Puerto-Rico, Canada, Germany, Outlying-US(Guam-USVI-etc), India, Japan, Greece, South, China, Cuba, Iran, Honduras, Philippines, Italy, Poland, Jamaica, Vietnam, Mexico, Portugal, Ireland, France, Dominican-Republic, Laos, Ecuador, Taiwan, Haiti, Columbia, Hungary, Guatemala, Nicaragua, Scotland, Thailand, Yugoslavia, El-Salvador, Trinadad&Tobago, Peru, Hong, Holand-Netherlands. |
| | Education-Num | Bachelors, Some-college, 11th, HS-grad, Prof-school, Assoc-acdm, Assoc-voc, 9th, 7th-8th, 12th, Masters, 1st-4th, 10th, Doctorate, 5th-6th, Preschool. |
| | Marital Status | Married-civ-spouse, Divorced, Never-married, Separated, Widowed, Married-spouse-absent, Married-AF-spouse. |
| | Relationship | Wife, Own-child, Husband, Not-in-family, Other-relative, Unmarried. |
| | Race | White, Asian-Pac-Islander, Amer-Indian-Eskimo, Other, Black. |
| | Sex | Female, Male. |
| | Capital Gain | Continuous. |
| | Capital Loss | Continuous. |
| Guest Features | Workclass | Private, Self-emp-not-inc, Self-emp-inc, Federal-gov, Local-gov, State-gov, Without-pay, Never-worked. |
| | Occupation | Tech-support, Craft-repair, Other-service, Sales, Exec-managerial, Prof-specialty, Handlers-cleaners, Machine-op-inspct, Adm-clerical, Farming-fishing, Transport-moving, Priv-house-serv, Protective-serv, Armed-Forces. |
| | Hours per week | Continuous. |

Table 1: Census Income Data [Ron et al., 1996]. In our experiment, income is used as prediction target, and the last three work-related attributes are used as federated features from guest.

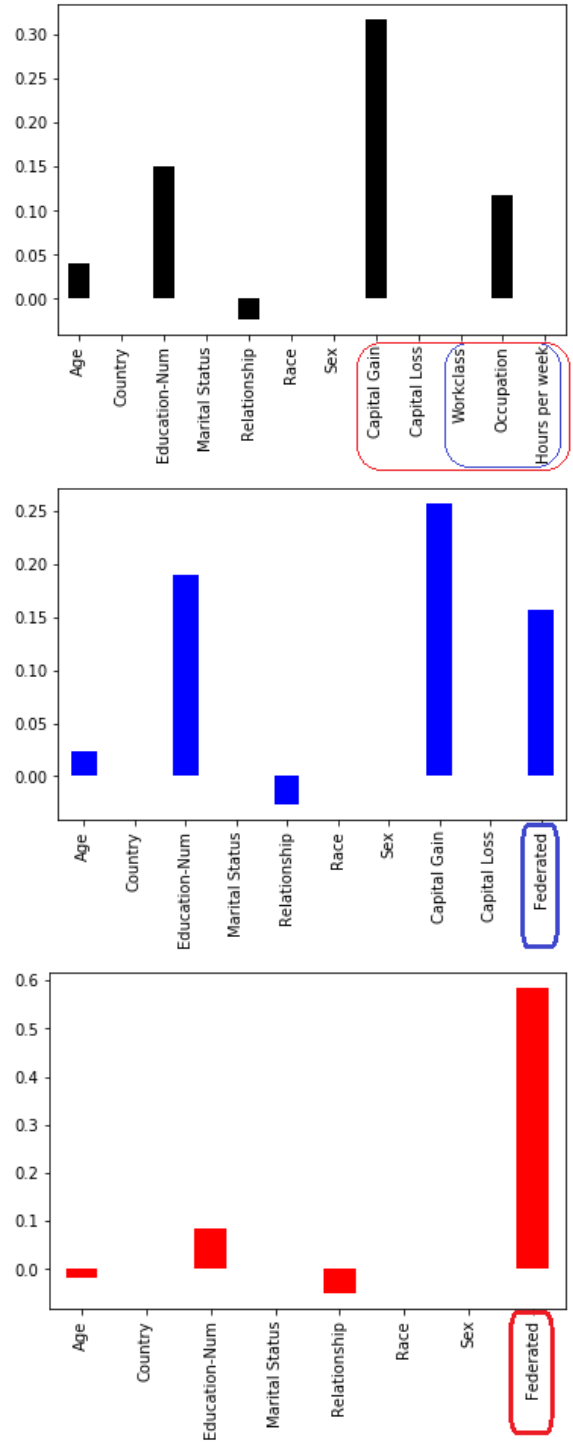

Figure 2: Feature Importance (Shapley values) for one single prediction. Top figure is for the whole feature space, middle figure is for federated feature of last 3 features, and bottom figure is for federated feature of last 5 features.

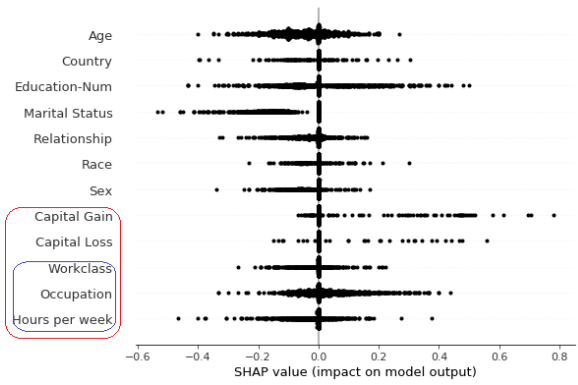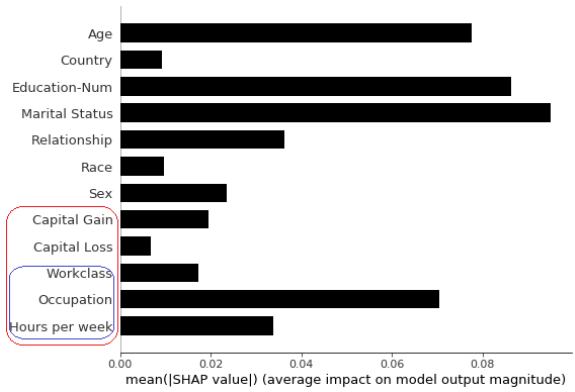
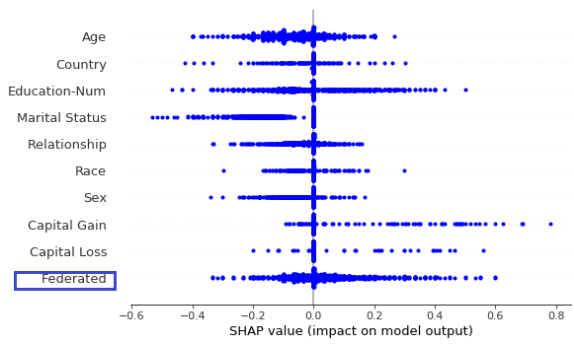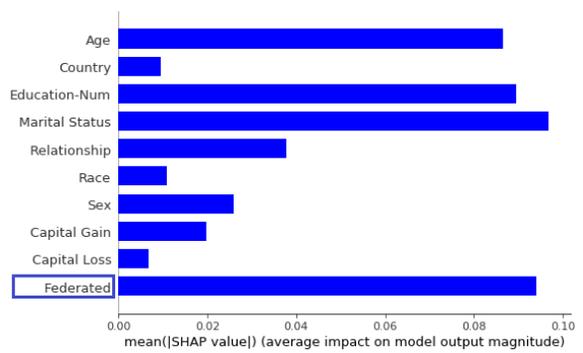
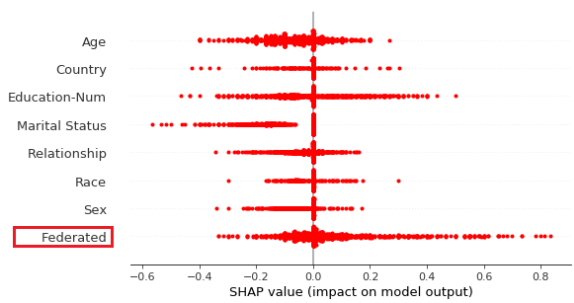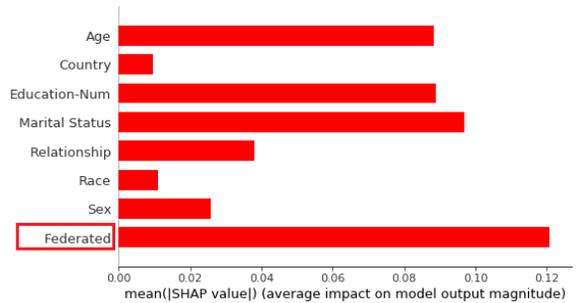

Figure 3: Scatter plot for Feature Importance (Shapley values) for 1000 random predictions. Top figure is for the whole feature space, middle figure is for federated feature of last 3 features, and bottom figure is for federated feature of last 5 features.

Figure 4: Bar plot for average Feature Importance (Shapley values) for 1000 random predictions. Top figure is for the whole feature space, middle figure is for federated feature of last 3 features, and bottom figure is for federated feature of last 5 features.